\documentclass{article}

\usepackage{arxiv}

\usepackage{etoolbox}
\makeatletter
\patchcmd{\@maketitle}{\@author}{\normalfont\@author}{}{}
\makeatother

\usepackage[utf8]{inputenc} 
\usepackage[T1]{fontenc}    
\usepackage{url}            
\usepackage{booktabs}       
\usepackage{amsfonts}       
\usepackage{nicefrac}       
\usepackage{microtype}      
\usepackage{lipsum}
\usepackage{graphicx}

\usepackage[hypertexnames=false, pdfborder={0 0 0}]{hyperref}
\usepackage{amsmath, amssymb, amsthm}
\usepackage{multirow}
\usepackage[table]{xcolor}
\usepackage{caption}
\usepackage{subfig}
\usepackage{algorithm}
\usepackage{algorithmic}
\usepackage{times}
\PassOptionsToPackage{numbers}{natbib}
\usepackage{natbib}  

\graphicspath{ {./images/} }

\title{Geometry Meets Semantics: Fractional Gradient Stabilization for Semantic-Driven Bounding Box Optimization in Visual Detection Tasks}
\author{
Qi Ming \\
Beijing University of Technology \\
chaser.ming@gmail.com
\and
Zheng Zhou \\
Beijing University of Technology \\
3613895315z@gmail.com
\and
Haitian Yang \\
ShanghaiTech University \\
yanght2023@shanghaitech.edu.cn
\and
Xudong Zhao \\
Beijing Institute of Technology \\
zhaoxudong@bit.edu.cn
\and
Mingjing Zhao \\
Beijing Electronic Science and Technology Institute \\
zmj@besti.edu.cn
\and
Liuqian Wang \\
Zhengzhou University \\
jeremy.wang0126@gmail.com
\and
Nanqing Liu \\
Yunnan Normal University \\
lansing163@163.com
}

\begin{document}
\maketitle
\begin{abstract}
Bounding boxes are fundamental for object localization in visual detection tasks. Among them, oriented bounding boxes are widely used in visual detection tasks, which provide a more precise directional representation. Generally, IoU-based losses are widely adopted to optimize box regression. However, we observed that IoU-driven box optimization suffers from two key issues: (1) it relies solely on geometric properties while ignoring semantic cues; (2) orientation optimization suffers from unstable gradients, causing oscillations in orientation convergence. In this paper, we propose a Fractional Semantic IoU loss to achieve unified semantic-geometric learning with gradient stabilization. First, we design a semantic similarity metric to guide IoU optimization, building a Semantic IoU loss (SIoU loss) with an adaptive gradient gating mechanism. Then, we revisit the gradient instability issue in oriented box optimization and extend the SIoU loss to a fractional-order formulation to build the \textbf{Fr}actional \textbf{S}emantic \textbf{IoU} \textbf{loss} (FrSIoU loss). The FrSIoU loss accumulates historical IoU states to regularize abnormal gradients during bounding box optimization process. Extensive experiments demonstrate that our approach achieves stable performance gains across different bounding box formulations and diverse visual detection tasks. The code will be available on GitHub.
\end{abstract}

\section{Introduction}
\label{sec:intro}
Visual detection serves as a fundamental task in computer vision, underpinning a wide range of downstream applications such as autonomous driving~\cite{zheng2020rotation,ming2023deep}, remote sensing analysis~\cite{yang2021learning,ming2024gradient,yang2021rethinking}, and robotic interaction~\cite{cheng2021grasp}. Furthermore, its success significantly impacts higher-level tasks, including tracking~\cite{yu2025point2rbox}, segmentation~\cite{ming2024towards}, and scene understanding~\cite{zuo2025fmgs}.

Bounding boxes are the most common geometric representation to localize object instances in detection frameworks. Depending on task requirements, bounding boxes can be categorized into Horizontal Bounding Boxes (HBBs)~\cite{ren2016faster,liu2016ssd}, Oriented Bounding Boxes (OBBs)~\cite{han2021align,ding2019learning}, polygon-based representations~\cite{yang2022Polygon}, etc. To achieve accurate bounding box optimization, modern detectors typically rely on regression losses such as L$_1$ loss, Smooth-L$_1$ loss~\cite{ren2016faster,liu2016ssd}, or Intersection-over-Union (IoU) based loss functions~\cite{zhou2019iou,rezatofighi2019generalized,zheng2020distance,ming2024gradient}. Among them, IoU-based losses have become a popular metric due to the scale-invariance of the metric and its direct reflection of localization quality. On this basis, a series of IoU-derived losses (e.g., GIoU~\cite{rezatofighi2019generalized}, DIoU~\cite{zheng2020distance},  EIoU~\cite{peng2021systematic}, GCIoU~\cite{ming2024gradient}) have been proposed, which improve regression stability and convergence by incorporating geometric factors such as distance, aspect ratio, or overlap consistency into the optimization process.

In recent years, OBB-based visual detection has been widely applied in various scenarios~\cite{zheng2020rotation,ming2023deep,yang2021learning,ming2024gradient,yang2021rethinking,cheng2021grasp}. As a generalized form of HBBs, OBBs explicitly encode object orientation, which is critical for high-precision detection and recognizing objects with large aspect ratios. For example, in remote sensing imagery, OBBs are commonly used to represent arbitrarily oriented objects such as bridges, ships~\cite{xia2018dota,li2020object}; in UAV aerial imagery, OBBs help distinguish densely packed vehicles~\cite{li2023developing,chen2024weakly}; and in 3D vision, voxel bounding boxes can be regarded as 2D OBBs extended with height information~\cite{zhou2019iou,sheng2022rethinking}. To achieve efficient OBB prediction and optimization, IoU losses have been extended to skew IoU formulations, achieving notable success in both 2D and 3D detection tasks~\cite{ming2021dynamic,zhou2019iou,sheng2022rethinking,ming2023deep,zheng2020rotation}.

\begin{figure}[t]
	\centering
\subfloat[Region-based IoU.]{
	\label{IoU}
	\includegraphics[width=0.22\textwidth]{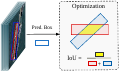}}\hspace{2.mm}
\subfloat[Pixel-based IoU.]{
	\label{PIoU}
	\includegraphics[width=0.22\textwidth]{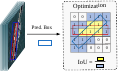}}\hspace{2.mm}
\subfloat[Our FrSIoU.]{
	\label{FrSIoU}
	\includegraphics[width=0.22\textwidth]{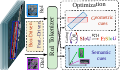}}\hspace{2.mm}
\subfloat[Loss gradient curve.]{
	\label{Grad}
	\includegraphics[width=0.22\textwidth,height=0.10\textheight]{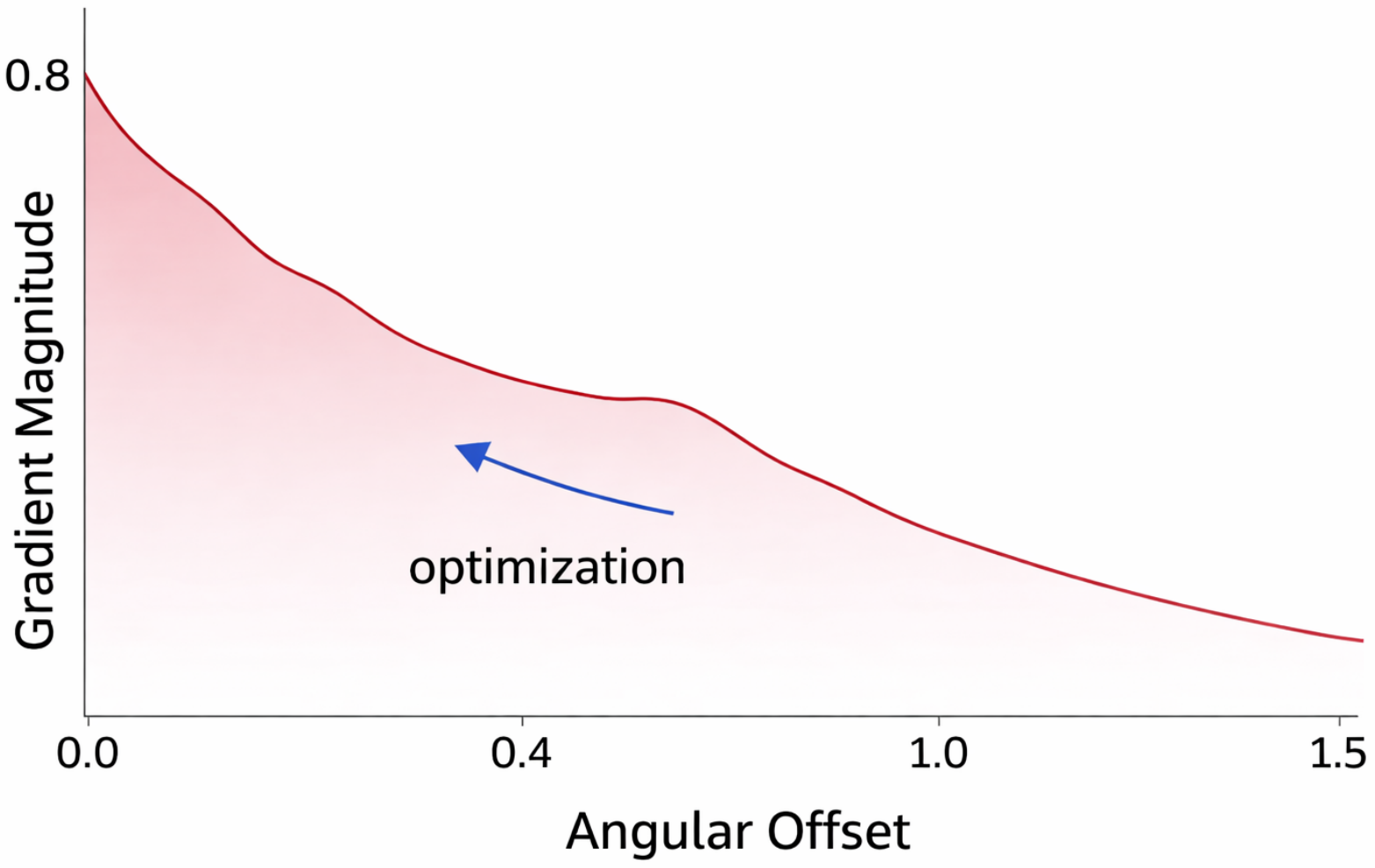}}
\caption{Calculation of different IoU variants (a-c) and visualization of abnormal gradients in IoU-based losses (d). Existing (a) region-based and (b) pixel-based IoUs consider only geometric constraints, while our FrSIoU (c) further incorporates semantic cues into box optimization.}
\end{figure}

However, existing methods remain constrained to the geometric properties of IoU when designing loss functions, which would leave their inherent limitations unresolved. We suggest that the IoU loss design should address two critical issues: (1)~\textbf{Current IoU-based losses primarily consider geometric constraints while ignoring the semantic content within bounding boxes}. As illustrated in Fig.~\ref{IoU} and Fig.~\ref{PIoU}, both region-based methods (e.g., GIoU~\cite{rezatofighi2019generalized}, DIoU~\cite{zheng2020distance}) and pixel-based methods (e.g., PIoU~\cite{chen2020piou}, PPIoU~\cite{endo2025polygon}) supervise box optimization purely from geometry with some constraints.  Intuitively, features within the box provide explicit semantic cues that could enhance localization accuracy. However, existing methods have not fully exploited this semantic information. (2)~\textbf{Skew IoU losses suffer from abnormal angular gradient oscillation during box optimization.} Ideally, with the model converging, gradients should diminish to allow fine-grained refinement. However, as shown in Fig.~\ref{Grad}, the optimization process exhibits unexpected gradient surges, which have also been discussed in \cite{ming2024gradient}. The abnormal gradients lead to oscillations in box regression and hindering high-precision.

Based on the above analysis, we propose the fractional semantic IoU loss to achieve unified semantic-geometric learning and to mitigate abnormal gradient variations for precise OBB optimization in challenging visual detection tasks. First, as illustrated in Fig.~\ref{FrSIoU}, we introduce a semantic similarity metric that extends IoU loss into Semantic IoU loss  (SIoU loss). SIoU loss jointly considers both geometric properties and semantic content of bounding boxes, while a gating mechanism dynamically balances the contributions of semantic and geometric similarities in bounding box optimization. Without the carefully designed geometric penalty terms (such as those in GIoU~\cite{rezatofighi2019generalized}, DIoU~\cite{zheng2020distance}, WIoU~\cite{tong2023wise}, etc.), SIoU loss naturally addresses common issues in naive IoU optimization, such as the non-overlapping regression problem~\cite{rezatofighi2019generalized}, achieving more efficient bounding box optimization. To further address gradient instability, we extend SIoU loss into a fractional formulation and develop the Fractional Semantic IoU loss (FrSIoU loss), which stabilizes angular gradients during model convergence and ensures accurate OBB regression. FrSIoU loss uses the historical semantic-geometric similarity to generate modulation weights dynamically, and regularize unstable angular gradients during the optimization. Our main contributions are summarized as follows:

\begin{itemize}
    \item We revisit the limitation of current IoU losses and conclude that existing methods rely solely on geometric constraints. Then, a Semantic IoU loss (SIoU loss) is proposed for unified semantic-geometric learning. SIoU loss incorporates both geometric and semantic consistency via a dynamic gating mechanism to adaptively guide bounding box regression.
    \item To address gradient instability in OBB optimization, we further extend SIoU loss into a Fractional Semantic IoU loss (FrSIoU loss). FrSIoU loss stabilizes model convergence via introducing a fractional derivative to capture the historical sIoU, which dynamically adjusts the loss weights  to regularize unstable angular gradients.
    \item We conduct extensive evaluations across multiple detectors and vision detection tasks. Experimental results demonstrate that FrSIoU loss exhibits strong generalization ability, consistently achieving performance gains and offering new insights for future research.
\end{itemize}

\section{Related Work}
\label{sec:formatting}

\paragraph{OBB Detection in Visual Detection Tasks.}
Oriented bounding boxes (OBBs) generalize horizontal bounding boxes (HBBs) by explicitly modeling object orientation, enabling high-precision detection of arbitrary-oriented objects. They have been widely applied across diverse vision tasks~\cite{cheng2021grasp,xia2018dota,chen2024weakly,ming2023deep,shi2018real,endo2025polygon,zhu2025five}. For example, OBBs are used to represent grasp poses in robotic manipulation~\cite{cheng2021grasp}, detect oriented objects in remote sensing benchmarks~\cite{xia2018dota}, and distinguish densely arranged vehicles in UAV imagery~\cite{chen2024weakly}. In autonomous driving, 3D bounding boxes can be viewed as OBBs with an additional height dimension~\cite{ming2023deep,sheng2022rethinking}. They are also adopted in face detection, scene text detection, and precision agriculture~\cite{shi2018real,endo2025polygon,liao2018rotation,zhu2025five}. The key advantage of OBBs lies in their generality, as they unify both axis-aligned and oriented representations, with HBB being a special case. Other representations, such as point sets~\cite{li2022oriented,song2023optimized}, Gaussian models~\cite{yang2021learning,yang2021rethinking}, polygons~\cite{yang2022Polygon,endo2025polygon}, and vector encodings~\cite{yi2021oriented}, can often be transformed into or approximated by OBBs. This highlights OBBs as a unified, flexible, and fundamental representation for  visual detection tasks.

\paragraph{IoU-based Losses for Bounding Box Optimization.}
Intersection-over-Union (IoU) is the dominant metric for object detection due to its scale-invariance and intuitive geometric interpretation, and IoU-based losses align training objectives with evaluation~\cite{zhou2019iou}. However, they suffer from slow convergence and zero gradients when predicted and ground-truth boxes do not overlap. To mitigate these issues, many variants introduce auxiliary geometric constraints: GIoU~\cite{rezatofighi2019generalized} adds an enclosing-box penalty to provide gradients in non-overlapping cases, DIoU and CIoU~\cite{zheng2020distance} incorporate center distance and aspect ratio consistency to accelerate convergence, and WIoU~\cite{tong2023wise} adaptively reweights samples to improve robustness against low-quality predictions. Beyond geometric extensions, alternative formulations approximate IoU by modeling boxes as distributions, such as Gaussian Wasserstein Distance (GWD)~\cite{yang2021rethinking}, Kullback-Leibler Divergence (KLD)~\cite{yang2021learning}, and Kalman filter-based KFIoU~\cite{yang2022kfiou}. While effective, these approaches still rely on geometric or surrogate supervision, which may partially break the consistency between training loss and evaluation metric. More recently, works have investigated the fundamental gradient behavior of IoU-based losses, revealing vanishing and oscillatory gradients near convergence for HBB~\cite{peng2021systematic}, and gradient anomalies with respect to angle and scale in OBB regression~\cite{ming2024gradient}, highlighting the need for more stable and well-behaved optimization strategies.

\section{Methodology}

\subsection{Problem Formulation}

\begin{figure}[t]
    \centering
    \includegraphics[width=0.95\textwidth]{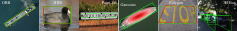}
    \caption{Illustration of different types of bounding box representations. OBB shows strong generality and can be widely applied across various visual detection tasks.}
    \label{presentation}
\end{figure}

\noindent\textbf{Unified OBB Representations}
Most visual detection tasks represent object instances by encoding object position, scale, and orientation.  The 2D OBB is commonly defined as $ \mathcal{B}_{\mathrm{OBB}}=(x,y,w,h,\theta) $, where $(x,y)\in\mathbb{R}^2$ denotes the box center, $w,h$ denote the box extents along the local axes, and $\theta\in\mathbb{R}$ is the rotation angle. We usually normalize the angle to a canonical interval, e.g, $\theta\in[-\tfrac{\pi}{2},\tfrac{\pi}{2})$ or $\theta\in[0,\pi)$ depending on the application~\cite{han2021align,ding2019learning}. Next, we adopt a unified view of these representations and their conversions for analyzing optimization behaviour.
(1) \textit{HBB $\leftrightarrow $ OBB.}  The axis-aligned HBB is widely used in general object detection tasks~\cite{lin2014microsoft,everingham2010pascal,ren2016faster,redmon2016you}. HBB is a special case of the OBB without orientation information: $ \mathcal{B}_{\mathrm{HBB}}=(x,y,w,h) $ and the corresponding OBB format is $ \mathcal{B}_{\mathrm{OBB}}=(x,y,w,h,0) $.
(2)\textit{Point-set $\leftrightarrow$ OBB.}  A point-set representation describes an object by a finite set of 2D points $ \mathcal{P}=\{p_i=(x_i,y_i)\}_{i=1}^N $~\cite{li2022oriented,song2023optimized}. Its equivalent OBB can be obtained by computing the minimum-area enclosing rectangle of $ \mathcal{P} $ (or of its convex hull). For complex polygons, PCA is the commonly used approximation to obtain the orthogonal direction.
(3) \textit{Gaussian representation $\leftrightarrow $ OBB.}
2D OBB can be viewed as a Gaussian distribution by mean and covariance, $ \mathcal{G}=(\mu,\Sigma) $. More detailed transformation process could be found in~\cite{yang2021learning,yang2021rethinking}.
(4) \textit{Polygon $\leftrightarrow$ OBB.}
Given a polygon with ordered vertices $\{v_i\}$, the minimum-area bounding OBB can be computed exactly via the rotating-calipers algorithm~\cite{shamos1978computational} applied to the convex hull. We denote this operator by $ \mathcal{B}_{\mathrm{OBB}}=\operatorname{MinAreaRect}(\operatorname{ConvexHull}(\{v_i\})) $.
(5) \textit{3D bounding box $\leftrightarrow$  OBB.}
A 3D box $ \mathcal{B}_{\mathrm{3D}}=(x,y,z,l,w,h,\phi) $ can be regarded as a 2D OBB with height, since it only has one rotational degree of freedom (yaw). Thus, many 2D OBB regression methods extend directly to 3D detection~\cite{ming2023deep,sheng2022rethinking}. As shown in Fig.~\ref{presentation}, most representations can be mapped to a unified OBB parametrization, enabling a common optimization perspective across diverse detection formulations.




\noindent \textbf{Anomalous Gradient Analysis for OBB Optimization}
To ensure efficient and stable OBB optimization, the skew IoU loss is widely employed in visual detection tasks~\cite{zhou2019iou,ming2024gradient,ming2023deep}.  Given the predicted box $\mathbf{b}_1=(x_1,y_1,w_1,h_1,\theta_1)$ and the corresponding target box $\mathbf{b}_2=(x_2,y_2,w_2,h_2,\theta_2)$, the angular deviation is $\theta=|\theta_1-\theta_2|$.  The classic linear IoU loss is

\begin{equation}
    \label{eq:linear_iou_loss}
    L_{\mathrm{IoU}}(\mathbf{b}_1,\mathbf{b}_2) \;=\; 1 - \mathrm{IoU}(\mathbf{b}_1,\mathbf{b}_2).
\end{equation}

Previous studies~\cite{ming2024gradient} have shown that center point convergence in OBB regression is fast and stable, and therefore, the skew IoU computation can be simplified to center-aligned OBBs. In this case, we obtain the loss gradient with respect to  $\theta$ as follows:

\begin{equation}
    \frac{\partial L_{\mathrm{IoU}}}{\partial\theta}
    = \mathrm{IoU}(\theta)\,\bigl(1+\mathrm{IoU}(\theta)\bigr)\,\cot\theta\;.
\label{eq:dL_linear}
\end{equation}
The detailed proof can be found in the Appendix. Eq.(\ref{eq:dL_linear}) reveals two mechanisms that produce unstable angular gradients in IoU-driven optimization:
\begin{enumerate}
  \item \emph{Singularity as $\theta\to 0$} : As the model converges, $\theta\downarrow 0$ and therefore $\cot\theta\to\infty$. The angular derivative grows as the angular error decreases, which produces large gradient magnitudes and leads to loss oscillation in the late stage of optimization. Due to boundary conditions in the skew IoU computation, $\theta$ never reaches $0$ (see Appendix for proof). Thus, the IoU loss remains convergent. However, its gradient variation during convergence still impedes bounding box refinement.
  \item \emph{Dependence on overlap magnitude}: The linear IoU loss gradient contains the multiplicative factor $\mathrm{IoU}(1+\mathrm{IoU})$, which can increase with IoU and therefore amplify the singular behaviour.
\end{enumerate}

These observations motivate two design goals for a robust localization loss: (1) incorporate semantic cues inside the region to complement purely geometric supervision, and (2) introduce mechanisms to stabilize angular gradients during optimization. In the following sections, we present our semantics-driven bounding box optimization and the fractional IoU loss to achieve the goals.

\subsection{Semantics-Driven Bounding Box Optimization}
\label{sec:siou}
Most IoU-based losses focus exclusively on geometric constraints and neglect the rich semantic cues embedded in deep features.
In this section, we will introduce a semantics-aware coupling that complements geometric cues with feature similarity, enabling continuous and adaptive guidance throughout the bounding box optimization process.

\noindent\textbf{Semantic score construction}. Let \(\mathbf{b}=(x,y,w,h,\theta)\) and \(\mathbf{b}^*=(x^*,y^*,w^*,h^*,\theta^*)\) denote a predicted OBB and a ground-truth OBB, and \(\mathcal{F}(\cdot)\) be the feature map extracted via the backbone network. We extract two semantic cues:
\begin{itemize}
    \item \emph{Intra-region feature distance.}
    Sample \(N\) paired locations inside the predicted and GT boxes (after appropriate warping such as RoI Align~\cite{he2017mask}), yielding feature vectors \(\{f_p^{(i)}\}_{i=1}^N\) and \(\{f_g^{(i)}\}_{i=1}^N\).
    Define the mean squared feature distance:
    \begin{equation}
      d_{\text{intra}} = \frac{1}{N}\sum_{i=1}^N \big\|f_p^{(i)} - f_g^{(i)}\big\|_2^2.
    \end{equation}
    Convert distance to a similarity \(s_1\in(0,1]\) via:
    \begin{equation}
      s_1 = \exp\big(-\alpha\, d_{\text{intra}}\big),
    \end{equation}
    where \(\alpha>0\) controls sensitivity.

    \item \emph{Context-region similarity.}
    Build horizontally-aligned context boxes \(\hat{\mathbf{b}}\) and \(\hat{\mathbf{b}}^*\) (e.g., axis-aligned expansions of \(\mathbf{b}\) and \(\mathbf{b}^*\)) and  extract their corresponding feature vectors $\{\hat{f}_p^{(i)}\}_{i=1}^{M}$ and $\{\hat{f}_g^{(i)}\}_{i=1}^{M}$ and compute  cosine similarity to obtain a contextual alignment score $s_2$:
    \begin{equation}
        s_2 = \frac{1}{2}\left(\frac{1}{M} \sum_{i=1}^{M}
    \frac{\hat{f}_p^{(i)} \cdot \hat{f}_g^{(i)}}{\|\hat{f}_p^{(i)}\|_2 \|\hat{f}_g^{(i)}\|_2}+1\right).
    \end{equation}
  \end{itemize}
Finally, the overall semantic score $s$ is defined as:
\begin{equation}
    \label{eq::s}
  s = \beta s_1 + (1-\beta) s_2,
\end{equation}
where $\beta$ is generally set to 0.5. It controls the balance between fine-grained object semantics and contextual semantics, so that the final semantic score \(s\in[0,1]\)

In Eq.(\ref{eq::s}), $s_1$ reflects the fine-grained semantic consistency between predicted and GT regions, providing a precise, discriminative metric that captures localized appearance similarity. Intuitively, the supervision from $s_1$ implicitly encodes geometric constraints between predicted boxes and GT objects. For example, if $s_1 \rightarrow 1$, the corresponding geometric IoU would naturally reach 1.

However, the precisely rotated RoI features would remove orientation cues, making it difficult for the model to converge with only $s_1$ supervision (see Appendix for detailed analysis). To address this issue, we introduce cosine similarity to compute $s_2$. Since contextual semantic information is more flexible, the correlations of context features help to optimize predicted RoI even without perfectly matching the GT region. Therefore, cosine similarity serves as a directional constraint to guide the bounding box optimization.

\noindent\textbf{Semantic IoU loss}.
Denote the geometric IoU by \(u=\mathrm{IoU}(\mathbf{b},\mathbf{b}^*)\in[0,1]\). We adopt the coupling
\begin{equation}
  \text{sIoU} \;=\; s + (1-s)\,u,
\end{equation}
which preserves the property that a perfect geometric overlap (\(u=1\)) yields \(\text{sIoU}=1\). Then, the complementary residual serves as
\begin{equation}
  L_{\mathrm{SIoU}} \;=\; (1-s)(1-u).
\end{equation}
$L_{\mathrm{SIoU}}$ is the semantics-driven IoU loss that measures the semantic-geometric misalignment. It extends the geometric-based IoU loss by introducing a feature-driven gating mechanism.  When $u=0$, the loss degenerates to $\mathcal{L}_{sIoU}=1-s$, providing non-zero gradients through the semantic term and thus avoiding the vanishing-gradient problem.  Therefore, $\mathcal{L}_{sIoU}$  enables feature-level supervision even under complete geometric mismatch.  Furthermore, by modulating the geometric term with the semantic score, $\mathcal{L}_{sIoU}$ adaptively emphasizes semantic or geometric alignment depending on the optimization phase.




\noindent\textbf{Analysis of Adaptive Gradient Gating}.
We next conduct the analysis of the gradient of the SIoU loss. Taking the bounding box parameter $\theta$ as an example, we obtain:
\begin{equation}
    \label{eq::gradsiou}
    \frac{\partial L_{\mathrm{SIoU}}}{\partial \theta}
    \;=\; - (1-s)\,\frac{\partial u}{\partial \theta}
          \;-\; (1-u)\,\frac{\partial s}{\partial \theta}.
  \end{equation}
Eq.(\ref{eq::gradsiou}) exposes the semantic-geometric gating mechanism in SIoU loss during training. When $s$ is large, semantic consistency becomes more confident, and semantic cues are expected to dominate the bounding box optimization. For example, as $s \to 1$, $(1 - s) \to 0$, and the overall gradient is dominated by the semantic term $-\partial s / \partial \theta$. Conversely, when $s$ is relatively small while $u$ is large, the semantic score is unreliable, and the model is expected to rely more on geometric cues for refinement. The semantic gradient $-\partial s / \partial \theta$ is scaled by $(1 - u)$; as $u \to 1$ and $s \to 1$, the semantic influence is suppressed, thereby  avoiding potential interference with  geometric alignment.

SIoU loss implements a soft arbitration between semantic and geometric cues. Unreliable semantics increase reliance on geometric information, while confident semantics suppress noisy geometric gradients. Through this adaptive gradient gating mechanism, the SIoU loss dynamically balances semantic and geometric supervision, enabling the network to select the dominant cue based on gradient flow for better bounding box regression.

\subsection{Fractional IoU Loss for Gradient Stabilization}
Though SIoU loss provides a flexible and adaptive bounding box optimization scheme, it still does not fundamentally resolve the issue of abnormal angular gradient variations. In this section, we will extend the SIoU loss into a fractional-order formulation to further stabilize abnormal gradients.

\noindent\textbf{Fractional IoU Loss}. Firstly, a fractional semantic IoU (FrSIoU) loss is proposed to incorporate a memory mechanism of the optimization process. It not only concerns the current loss, but also dynamically adjusts the loss function based on the historical semantic-geometric similarity. Then, it can effectively regularize unstable gradients.

The fractional derivative extends differential operations to non-integer orders and characterizes the nonlocal effects of network training~\cite{khalil2014new}. The Grünwald-Letnikov (G-L) approximation is used to compute the $v$-th derivative at the current epoch~\cite{scherer2011grunwald}. Given $M$ historical steps, the G-L derivative is defined as
\begin{equation}
D_t^v(sIoU) = \sum_{k=0}^{M} c_k \cdot sIoU_{t-k},
\end{equation}
where $c_k$ is calculated as
\begin{equation}
c_0 = 1,\quad \frac{k - 1 - v}{k} \cdot c_k = c_{k-1} \quad \text{for } k \ge 1,
\end{equation}

Taking $M=2$ as an example, the approximate expression is
\begin{equation}
\begin{array}{*{20}{l}}
{D_t^v (sIoU)}&{ \approx {c_0}sIo{U_t} + {c_1}sIo{U_{t - 1}} + {c_2}sIo{U_{t - 2}}}\\
{}&{ = sIo{U_t} - v sIo{U_{t - 1}} + \frac{{v (v  - 1)}}{2}sIo{U_{t - 2}}},
\end{array}
\end{equation}
The proposed derivative $D_t^\nu(sIoU)$ represents the trend of semantic-geometric similarity, which further leads to the definition of the modulation weight $W_t$ as
\begin{equation}
W_t = \exp(-\lambda \cdot D_t^v(sIoU)),
\end{equation}
where $\lambda$ is a hyperparameter controlling the modulation intensity. Finally, the fractional semantic IoU (FrSIoU) loss is defined as
\begin{equation}
\begin{array}{*{20}{l}}
{{{\cal L}_{FrSIoU}}}{ = {W_t} \cdot (1 - s)(1 - u)}.
\end{array}
\end{equation}

\noindent\textbf{Gradient Stabilization of Fractional IoU}. The fractional-order modulated weights $W_t$ make the optimization process more robust and stable by altering the gradient, while the fractional derivative $D_t^v(sIoU)$ quantifies the convergence tendency.

When the sIoU metric grows continuously $sIoU_t > sIoU_{t-1}$, given $c_0=1$ and $c_1=-v$, the derivative $D_t^v(sIoU) \approx sIoU_t - v*sIoU_{t-1}$ tend to be positive. Then, $W_t = \exp(-\lambda \cdot \text{positive value}) < 1$. Therefore, the loss is suppressed, which decays the optimization step size and mitigates the potential risk of the gradient explosion caused by the $cot \theta$ term when $\theta \to 0$. When the SIoU loss drops sharply, $D_t^v(sIoU)$ becomes negative, which results in $W_t = \exp(-\lambda \cdot \text{negative value}) > 1$. The loss is then amplified, providing a strong and effective correction signal and imposing corrective constraints.

The full gradient of the FrSIoU loss function with the angle parameter $\theta$ proves this gradient stabilization. 
\begin{equation}
\begin{array}{*{20}{l}}
{\frac{{\partial {{\cal L}_{FrSIoU}}}}{{\partial \theta }}}{ = \frac{{\partial ({W_t} \cdot {L_{base}})}}{{\partial \theta }}}
{}{ = \frac{{\partial {W_t}}}{{\partial \theta }}{L_{base}} + {W_t}\frac{{\partial {L_{base}}}}{{\partial \theta }}}
\end{array}
\end{equation}

In this part, a fundamental assumption is that the semantic score $s$ is treated as a constant whose partial derivative of the geometric parameter $\theta$ is zero ($\frac{\partial s}{\partial \theta} = 0$). This is because $s$ is extracted from deep features, whose gradient flow is separate from geometric regression. Then, the corresponding partial derivatives are further simplified as
\begin{equation}
\begin{array}{*{20}{l}}
{\frac{{\partial {L_{base}}}}{{\partial \theta }}}{ = \frac{{\partial ((1 - s)(1 - u))}}{{\partial \theta }}}
{}{ =  - (1 - {s_t})\frac{{\partial {u_t}}}{{\partial \theta }}}
\end{array}
\end{equation}

\begin{equation}
\begin{array}{*{20}{l}}
{\frac{{\partial sIo{U_{t - k}}}}{{\partial \theta }}}{ = \frac{{\partial ({s_{t - k}} + (1 - {s_{t - k}}){u_{t - k}})}}{{\partial \theta }}}
{}{ = (1 - {s_{t - k}})\frac{{\partial {u_{t - k}}}}{{\partial \theta }}}
\end{array}
\end{equation}

\begin{equation}
\begin{array}{*{20}{l}}
{\frac{{\partial D_t^v (sIoU)}}{{\partial \theta }}}{ = \sum\limits_{k = 0}^M {{c_k}} \frac{{\partial sIo{U_{t - k}}}}{{\partial \theta }}}
{}{ = \sum\limits_{k = 0}^M {{c_k}} (1 - {s_{t - k}})\frac{{\partial {u_{t - k}}}}{{\partial \theta }}}
\end{array}
\end{equation}

\begin{equation}
\begin{array}{*{20}{l}}
{\frac{{\partial {W_t}}}{{\partial \theta }}}{ = \frac{{\partial \exp ( - \lambda D_t^v)}}{{\partial \theta }}}
{}{ = {W_t} \cdot ( - \lambda ) \cdot \frac{{\partial D_t^v(sIoU)}}{{\partial \theta }}}
\end{array}
\end{equation}

Then, the full gradient expression is constructed as

\begin{equation}\label{eq:fiou}
\begin{array}{*{20}{l}}
{\frac{{\partial {{\cal L}_{FrSIoU}}}}{{\partial \theta }} = {W_t}\left( {\frac{{\partial {L_{base}}}}{{\partial \theta }} - \lambda {L_{base}}\frac{{\partial D_t^\nu (sIoU)}}{{\partial \theta }}} \right)}\\
{ = {W_t}\left[ { - (1 - {s_t})\frac{{\partial {u_t}}}{{\partial \theta }} - \lambda {L_{base}}\left( {\sum\limits_{k = 0}^M {{c_k}} (1 - {s_{t - k}})\frac{{\partial {u_{t - k}}}}{{\partial \theta }}} \right)} \right]}
\end{array}
\end{equation}
Eq. (\ref{eq:fiou}) demonstrates that the total gradient is a weighted sum of historical geometric gradients. The optimization process is regularized by the historical states, which reduces the gradient changes caused by the $\cot \theta$ term and achieves stable convergence for high-precision positioning.

\begin{table*}[!t]
  \renewcommand\arraystretch{1.2}
  \centering
  \caption{Ablation study of different loss functions on the challenging large-scale DOTA-v1.0 dataset based on Rotated RetinaNet.}
  \resizebox{0.99\textwidth}{!}{
  \begin{tabular}{c|ccccccccccccccc|c} 
  \toprule
  Loss terms  & PL & BD & BR & GTF & SV & LV & SH & TC & BC & ST & SBF & RA & HA & SP & HC & $\text{mAP}_{50}$(\%) \\ 
  \hline
  $L_1$ loss  & 87.2 & 75.2 & 34.8 & 63.5 & 66.4 & 75.4 & 85.0 & 90.3 & 61.5 & 60.5 & 52.4 & 62.2 & 60.0 & 56.8 & 50.4 & 65.44 \\
  GIoU loss & 89.1 & 77.1 & 37.2 & 67.8 & 67.9 & 74.3 & 85.1 & 90.5 & 64.3 & 60.4 & 48.7 & 63.1 & 61.7 & 56.5 & 56.7 & 66.67 \\
  SIoU loss & 89.2 & 77.3 & 37.4 & 67.0 & 67.8 & 74.5 & 85.4 & 90.5 & 61.9 & 64.3 & 49.8 & 65.1 & 62.0 & 55.4 & 54.6 & 66.82 \\ \hline
  \textbf{FrSIoU loss} & 89.2 & 76.9 & 37.6 & 67.6 & 67.9 & 74.6 & 85.5 & 90.5 & 63.5 & 64.1 & 48.8 & 65.2 & 62.0 & 56.6 & 57.6 & \textbf{67.27} \\
  \bottomrule
  \end{tabular}}
  \label{table:ablation_method}
  \end{table*}

\section{Experiments}

\subsection{Experimental Setup}

\noindent\textbf{Datasets}. We conduct experiments on multiple datasets to verify the generalization and robustness of our approach, including DOTA~\cite{xia2018dota}, HRSC2016~\cite{liu2017high} and SSDD~\cite{zhang2021ssdd}. DOTA-v1.0 is utilized as the primary benchmark to assess detection performance across diverse categories and complex scenes. As a large-scale aerial dataset, DOTA-v1.0 contains 2,800 images and 188,000 instances across 15 categories, characterized by extreme scale variations and arbitrary orientations. For fine-grained ship detection, HRSC2016 contains 1,061 images annotated with rotated boxes, specifically highlighting objects with large aspect ratios. In the SAR domain, we utilize the SSDD dataset to demonstrate the generalization capability and robustness of our approach across different imaging modalities and environmental conditions. The SSDD dataset provides 1,160 images for ship detection, focusing on detecting small objects within complex coastal environments.

\noindent\textbf{Evaluation Metrics}. Following common practice, we use three standard metrics to evaluate detection performance: mAP$_{50}$, mAP$_{75}$, and mAP. mAP$_{50}$ and mAP$_{75}$ represent the Average Precision at IoU thresholds of 0.5 and 0.75, respectively, while mAP averages the precision across IoU thresholds from 0.5 to 0.95 with a step of 0.05. Distinct from prior studies that primarily report mAP$_{50}$, we explicitly include mAP$_{75}$ and mAP to demonstrate the model's robustness in high-precision localization. This stricter evaluation protocol ensures a more rigorous comparison of bounding box regression accuracy.

\noindent\textbf{Implementation Details}. For fair comparison, all experiments are implemented based on    MMRotate framework~\cite{zhou2022mmrotate} using four NVIDIA GeForce RTX 3090 GPUs. We use the SGD optimizer with an initial learning rate of 0.02, momentum of 0.9, and weight decay of 0.0001. The batch size is set to 16 with a 500-iteration linear warm-up. Regarding training schedules, models on the DOTA dataset are trained for 24 epochs, while those on HRSC2016 and SSDD datasets are trained for 72 epochs. Specifically for the DOTA dataset, we adopt single-scale training and testing. Ablation studies are conducted on the validation set. For the main results, models are trained on the combined training and validation sets, and the final performance is obtained by submitting testing results to the official evaluation server.

\begin{table}[t]
	\centering
    \renewcommand\arraystretch{1.2}
    \setlength\tabcolsep{12.51pt}
    \small
    \caption{Analysis of hyperparameters in FrSIoU loss.}
    \resizebox{0.9\textwidth}{!}{
	\begin{tabular}{ccccccccc}
		\toprule
		$\lambda$     & $\alpha$   & $\text{mAP}_{50}$(\%)& $\lambda$   & $\alpha$   & $\text{mAP}_{50}$(\%) & 	$\lambda$   &  $\alpha$& $\text{mAP}_{50}$(\%)   \\
		\midrule
		\multirow{3}{*}{0.05} & 0.05 &67.21& \multirow{3}{*}{0.1} &0.05& 67.15 & \multirow{3}{*}{0.3} & 0.05 & 66.97 \\
		& 0.1 & 66.80 &                    & 0.1 &67.06 &                    & 0.1 & 67.17 \\ 
& 0.3 & 67.04 &                    & 0.3 & 67.09 &                    & 0.3 & $\textbf{67.27}$ \\ 
\bottomrule
	\end{tabular}}
\label{hyper}
\end{table}

\begin{table*}[t]
  \centering
    \renewcommand\arraystretch{1.3}
    \setlength\tabcolsep{5.51pt}
    \fontsize{7.5pt}{8pt}\selectfont
    \caption{Comparison with state-of-the-arts on the DOTA-v1.0 dataset using single-scale training and testing. The input image size is $1024 \times 1024$. The \textbf{best} and \underline{second-best} results are highlighted in \textbf{bold} and \underline{underlined}.}
    \label{dotasota}
\resizebox{1.0\textwidth}{!}{
  \begin{tabular}{c|c|ccccccccccccccc|ccc} 
  \toprule
    & Methods     & PL    & BD    & BR     & GTF   & SV      & LV     & SH    & TC      & BC    & ST    & SBF    & RA    & HA     & SP    & HC     & $\text{mAP}_{50}$ & $\text{mAP}_{75}$ & mAP    \\ 
  \hline
  \multirow{6}{*}{\rotatebox{90}{One Stage}} 
  & R-RetinaNet~\cite{lin2017focal}    & \textbf{89.65} & 82.57 & 38.16  & 69.84 & 77.41   & 62.76  & 77.25 & 90.69   & 83.80 & 82.04 & 59.93  & 64.83 & 57.38  & 64.76 & 45.56  & 69.77 & 37.70 & 39.65 \\
     & CSL~\cite{yang2020arbitrary}       & 89.34 & 79.67 & 40.99  & 69.96 & 77.71   & 62.08  & 77.51 & 90.87   & 82.87 & 82.04   & 60.03 & 65.27  & 53.60 & 64.08  & 46.61 & 69.51 & 40.41 & 39.69 \\

     & R$^3$Det~\cite{yang2021r3det}   & 89.29 & 75.21 & 45.42 & 69.24 & 75.54 & 72.90 & 79.29 & \underline{90.89} & 81.02 & 83.25 & 58.82 & 63.17 & 63.40 & 62.22 & 37.41 & 69.80 & 36.58 & 37.82 \\
     
      & GWD~\cite{yang2021rethinking}       & 88.93 & 77.04 & 45.85  & 69.30 & 72.53   & 64.05  & 76.41 & 90.87  & 79.19 & 80.46 & 57.68  & 64.37 & 63.60  & 64.75 & 48.25  & 69.55 & 38.91 & 39.50 \\
      & KLD~\cite{yang2021learning}   & 89.21 & 80.68 & 48.79  & 70.56 & 77.72   & 75.77  & 86.34 & \textbf{90.90}   & 82.83 & 83.78 & 65.70  & 65.92 & 66.08  & 66.42 & 40.70  & 72.76 & 37.83 & 39.92  \\
      & S$^2$A-Net~\cite{han2021align} & 89.30 & 80.51 & 50.38 & 73.23 & 78.41 & 77.40 & 86.80 & \underline{90.89} & 85.64 & 84.23 & 62.13 & 65.95 & 66.61 & 67.74 & 53.51 & 74.18 & 36.83 & 39.84 \\
  \hline
  \multirow{10}{*}{\rotatebox{90}{Two Stage}} 
     & RoI-Trans.~\cite{ding2019learning} & 88.97 & 82.12 & \underline{54.57} & \textbf{76.27} & 79.29 & 77.96 & 87.94 & \textbf{90.90} & \textbf{87.19} & 85.66 & 62.18 & 62.65 & 74.63 & \underline{72.43} & 59.23 & 76.13 & 46.88 & 45.56 \\

     & KFIoU~\cite{yang2022kfiou} & 89.05 & 75.17 & 49.04 & 69.67 & 78.06 & 75.45 & 86.69 & \textbf{90.90} & 83.65 & 84.48 & 62.21 & 62.85 & 66.72 & 65.95 & 50.20 & 72.67 & 36.01 & 38.88 \\
     
     & ReDet~\cite{han2021redet}     & 89.20 & \underline{83.78} & 52.19  & 71.05 & 78.05   & 82.50  & 88.24 & 90.86   & 87.26 & 85.97 & \textbf{65.50}  & 62.86 & \underline{75.89}  & 70.05 & \textbf{66.71}  & \underline{76.67} & 48.84 & 46.71 \\
     & O-Former~\cite{zhao2024orientedformer} & 88.14 & 79.13 & 51.96 & 67.34 & \textbf{81.02} & \textbf{83.26} & \textbf{88.29} & \textbf{90.90} & 85.57 & \underline{86.25}& 60.84 & \underline{66.36} & 73.81 & 71.23 & 56.49 & 75.37 & 46.39 & 45.01 \\

    & ACM~\cite{xu2024acm} & 88.06 & 73.99 & 49.08 & 61.54 & 79.73 & 77.23 & 87.07 & \underline{90.89} & 82.17 & 84.04 & 56.97 & 61.21 & 63.91 & 67.78 & 36.16 & 70.66 & 42.64 & 41.26 \\
    
     & Point2RBoxv2~\cite{yu2025point2rbox} & 88.21 & 69.76 & 23.30 & 44.22 & 79.82 & 76.18 & 87.09 & 89.27 & 44.97 & 83.26 & 16.19 & 41.00 & 44.95 & 58.00 & 49.54 & 59.72 & 25.96 & 30.13 \\


     & GSDet~\cite{ding2025gsdet}   & 88.50 & 79.02 & 50.58 & 71.71 & 78.53 & \underline{83.21} & \underline{88.25} & 90.84 & 84.26 & 80.33 & 55.87 & 57.04 & 74.70 & 67.87 & 57.46 & 73.88 & \textbf{52.17}& \underline{47.69} \\

     & GauCho~\cite{marques2025gaucho}   & 88.56 & 76.69 & 49.87 & 61.59 & 80.00 & 79.59 & 87.46 & 90.88 & 83.24 & 84.19 & 54.98 & 64.86 & 64.12 & 69.53 & 51.33 & 72.46 & 37.03 & 40.46 \\


    \cline{2-20}
     & O-RCNN~\cite{xie2021oriented}     & \underline{89.35} & 81.41 & 52.71 & \underline{75.02} & 79.03 & 82.43 & 87.82 & \underline{90.89} & \underline{86.40} & 85.30 & 63.37 & 65.69 & 68.28 & 70.48 & 57.22 & 75.69 & 47.18 & 45.19  \\ 
     & \textbf{FrSIoUloss (Ours)}   & 89.16 & \textbf{84.66} & \textbf{57.34} & 71.42 & \underline{80.11} & 82.62 & 87.98 & 90.84 & 80.56 & \textbf{86.66} & \underline{64.75} & \textbf{70.68} & \textbf{77.76}& \textbf{73.47} & \underline{61.30} & \textbf{77.28} & \underline{50.50} & \textbf{47.94} \\
   
  \bottomrule
  \end{tabular}
  }
\end{table*}

\subsection{Ablation Study and Analysis}


\noindent\textbf{Effect of Semantic IoU Loss}. We first evaluate the impact of introducing semantic guidance into box regression. The baseline model trained with standard $L_1$ loss achieves 65.44\% mAP. While the geometric-based GIoU loss improves this to 66.77\% by addressing the non-overlapping issue, it still lacks semantic awareness. By replacing GIoU with our proposed SIoU loss, the performance further improves to 66.82\% mAP.
Notably, SIoU loss demonstrates superior efficacy in rotation-sensitive categories. For instance, the accuracy for roundabout (RA) increases by 2.0\% compared to GIoU from 63.1\% to 65.1\%.
This improvement indicates that the semantic alignment term successfully constrains the optimization direction, enabling more robust regression for objects with complex appearances where purely geometric constraints may fall short.


\noindent\textbf{Effect of Fractional SIoU Loss}. Although SIoU loss enhances bounding box optimization flexibility, it does not fundamentally address the problem of angle gradient discontinuity. 
To further stabilize training, we introduce a Fractional SIoU loss that adaptively decomposes the angle term into periodic components, ensuring smooth gradient propagation near angle boundaries. 
As presented in Table~\ref{table:ablation_method}, FrSIoU achieves the best overall performance of 67.27\% mAP, outperforming the SIoU loss by 0.45\% and the $L_1$ baseline by 1.83\%.
Specific gains in categories such as bridge (BR) and small vehicle (SV) confirm that the fractional modulation effectively stabilizes angular gradients, ensuring smooth convergence and high-precision localization.

\noindent\textbf{Effect of Hyperparameters in FrSIoU loss}. We further analyze the impact of the modulation intensity $\lambda$ and the semantic sensitivity parameter $\alpha$ on detection performance, as reported in Table~\ref{hyper}. The results demonstrate that FrSIoU is highly robust to hyperparameter variations, with $\text{mAP}_{50}$ fluctuations remaining within a narrow range of $0.5\%$. This stability implies that the proposed FrSIoU effectively regularizes the gradient without requiring delicate tuning. Based on the peak $\text{mAP}_{50}$ of $67.27\%$, we set $\lambda=0.3$ and $\alpha=0.3$ as the standard configuration.

\begin{table}[t] 
\renewcommand\arraystretch{1.2}
  \centering
  \caption{Comparison of different models on HRSC2016 and SSDD datasets.}
    \setlength\tabcolsep{10.51pt}
    \label{tab:hrsc}
  \resizebox{0.9\textwidth}{!}{
    \begin{tabular}{cccccc}
      \toprule
      \multirow{2}{*}{\textbf{Model}} & \multicolumn{3}{c}{\textbf{HRSC2016}} & \multicolumn{1}{c}{\textbf{SSDD}} \\
      \cmidrule(lr){2-4} \cmidrule(lr){5-5}
       & mAP(\%)  & mAP$_{50}$(\%) & mAP$_{75}$(\%) & mAP$_{50}$(\%)  \\
      \midrule
      R-RetinaNet       & 46.63 & 81.60 & 48.80 & 78.32 \\
      R-RetinaNet \textbf{+  FrSIoU loss}  & 52.80 (+ 6.17)  & 83.00 (+1.40) & 59.30 (+10.50) & 80.49 (+2.17)  \\
      \midrule
      O-RCNN         & 65.81 & 89.30  & 79.70 & 89.45  \\
      O-RCNN \textbf{+ FrSIoU loss}    & 66.26 (+ 0.45) & 90.50 (+ 1.20) & 81.00 (+1.30)  & 90.07 (+ 0.62)  \\
      \bottomrule
    \end{tabular}
  }
\end{table}

\subsection{Results and Comparisons}
\noindent\textbf{Comparison with SOTA Methods}. We validate the proposed FrSIoU loss within the Oriented R-CNN~\cite{xie2021oriented} framework, chosen for its efficiency and strong baseline performance. Table~\ref{dotasota} compares our method with recent state-of-the-art oriented object detection approaches categorized into one-stage methods and two-stage methods. As presented in Table~\ref{dotasota}, our approach achieves the best overall performance among the compared methods, reaching an mAP of 47.94\% and mAP$_{50}$ of 77.28\%. Despite using a simple ResNet-50 backbone, our method outperforms recent competitive models, such as ReDet (46.71\% mAP) and GSDet (47.69\% mAP), which often rely on more complex feature refinement modules. 

A key advantage of FrSIoU loss is the ability to generate high-quality bounding boxes. 
Compared to the baseline Oriented R-CNN, our method yields a substantial improvement of 3.32\% in mAP$_{75}$ and 2.75\% in mAP. Notably, significant gains are observed in categories requiring precise orientation estimation, such as bridge (BR) (+4.63\%) and roundabout (RA) (+4.99\%). These results confirm that by stabilizing angular gradients during the later stages of training, FrSIoU loss effectively prevents oscillation and enables the model to satisfy stricter IoU thresholds, validating its effectiveness for high-precision detection tasks.

\noindent\textbf{Generalization on Ship Detection Datasets}. To verify the robustness of our approach across specific domains, we extend our evaluation to two ship detection benchmarks: the optical HRSC2016 dataset and the SAR-based SSDD dataset. As detailed in Table~\ref{tab:hrsc}, incorporating FrSIoU loss yields consistent performance gains across different detector architectures (Rotated RetinaNet and Oriented R-CNN) and imaging modalities. On the HRSC2016 dataset, our method boosts the overall mAP of Rotated RetinaNet by 6.17\% and achieves a 10.50\% increase in the stricter mAP$_{75}$ metric. This substantial improvement highlights the efficacy of the fractional stabilization mechanism in localizing elongated objects, where minor angular perturbations typically cause severe IoU degradation. Furthermore, on the SSDD dataset, FrSIoU improves the mAP$_{50}$ for both one-stage and two-stage detectors, confirming its adaptability to diverse sensing scenarios beyond optical imagery.

\section{Conclusion}


In this paper, we revisited the limitations of IoU-based bounding box optimization, identifying the key challenges of the absence of semantic supervision and unstable angular gradients in oriented bounding box regression. Then, we propose a Semantics IoU loss to utilize geometric constraints with adaptive feature-level guidance for better bounding box regression. Building on this, we introduce the Fractional Semantic IoU (FrSIoU) loss to  mitigate angular gradient singularities via  incorporating a fractional-order memory mechanism. Extensive experiments on OBB benchmarks demonstrate the superior performance and generalization capability on visual detection tasks. The significant gains in high-precision metrics across mainstream detectors confirm that FrSIoU loss provides a robust solution for precisely oriented object detection, offering a new direction for designing stable objectives in complex geometric regression tasks.

\newpage
\bibliographystyle{unsrt} 
\small
\bibliography{references}
\normalsize

\end{document}